\documentclass[conference,10pt]{IEEEtran}
\IEEEoverridecommandlockouts

\usepackage{cite}
\usepackage{amsmath,amssymb,amsfonts}
\usepackage{graphicx}
\usepackage{algorithmic}
\usepackage{textcomp}
\usepackage{xcolor}

\usepackage{ifpdf}
\usepackage{booktabs} 
\usepackage{booktabs}
\usepackage{subfigure}
\usepackage[linesnumbered]{algorithm2e}
\usepackage[T1]{fontenc}
\usepackage{mathrsfs}
\usepackage{amsmath,bm}
\usepackage{color}
\usepackage{authblk}
\usepackage{comment}

\def\BibTeX{{\rm B\kern-.05em{\sc i\kern-.025em b}\kern-.08em
    T\kern-.1667em\lower.7ex\hbox{E}\kern-.125emX}}

\begin{document}

\title{Invocation-driven Neural Approximate Computing with a Multiclass-Classifier and Multiple Approximators \thanks{This research was partially supported by National Natural Science Foundation of China (Grant No. 61602300, 61432017), 
Shanghai Science and Technology Committee (Grant No. 18ZR1421400),
Shanghai Jiao Tong University Biomedical Engineering Research Foundation (No. YG2015MS17), 
and Shanghai clinical ability construction of The three grade hospital (No. SHDC12015904). 
The Corresponding author is Li Jiang.
\authorcr
Permission to make digital or hard copies of all or part of this work for
personal or classroom use is granted without fee provided that copies are not
made or distributed for profit or commercial advantage and that copies bear
this notice and the full citation on the first page. Copyrights for components
of this work owned by others than ACM must be honored. Abstracting with
credit is permitted. To copy otherwise, or republish, to post on servers or to
redistribute to lists, requires prior specific permission and$/$or a fee. Request
permissions from Permissions$@$acm.org.
\authorcr
\emph{ICCAD} $'$18, November 5–8, 2018, San Diego, CA, USA
\authorcr
\textcopyright 2018 Association for Computing Machinery.
\authorcr
ACM ISBN 978-1-4503-5950-4/18/11...\$15.00
\authorcr
https://doi.org/10.1145/3240765.3240819
}}

\author[*]{Haiyue Song}
\author[*]{Chengwen Xu}
\author[**]{Qiang Xu}
\author[*]{Zhuoran Song}
\author[*]{Naifeng Jing}
\author[*]{Xiaoyao Liang}
\author[*]{Li Jiang}
\affil[*]{MoE Key Lab of Artificial Intelligence, AI Institute}
\affil[*]{School of Electronic, Information and Electrical Engineering, Shanghai Jiao Tong University, China
}
\affil[**]{Department of Computer Science and Engineering, The Chinese University of Hong Kong, Hong Kong}

\maketitle



\begin{abstract}

Neural approximate computing gains enormous energy-efficiency at the cost of tolerable quality-loss. 
A neural approximator can map the input data to output while a classifier determines whether the input data are safe to approximate with quality guarantee. However, existing works cannot maximize the invocation of the approximator, 
resulting in limited speedup and energy saving. By exploring the mapping space of those target functions, in this paper, we observe a nonuniform distribution of the approximation error incurred by the same approximator. We thus propose a novel approximate computing architecture with a Multiclass-Classifier and Multiple Approximators (MCMA). These approximators have identica network topologies, and thus can share the same hardware resource in an neural processing unit(NPU) clip. In the runtime, MCMA can swap in the invoked approximator by merely shipping the synapse weights from the on-chip memory to the buffers near MAC within a cycle. We also propose efficient co-training methods for such MCMA architecture. Experimental results show a more substantial invocation of MCMA as well as the gain of energy-efficiency.
\end{abstract}


\section{Introduction}


    Approximate computing is a promising technique to gain energy efficiency with tolerable losses of computation quality for specific error-tolerant applications, such as recognition, mining, and search (RMS) applications. Existing works strive to reduce the computation effort by precision scaling~\cite{Tian2015ApproxMA}, loop perforation~\cite{Sidiroglou2011Managing}, memorization~\cite{Rahimi2013Spatial}; others try to minimize the performance hurdle by skipping memory references~\cite{Samadi-sigarch14}. Approximate computing can also gain more parallelism using strategies like lowering the precision of intermediate computations~\cite{Roldaolopes2009More}, memorization~\cite{Sinha-tvlsi16}, etc. The reported power saving is usually within $5\% - 40\%$~\cite{esmaeilzadeh-asplos12}, depending on the application, the approximation technique, and the acceptable error. 

Neural networks (NNs) are suitable for general approximate computing. First, NNs are inherently utilized in many error-tolerant applications, such as real-world pattern recognition~\cite{Han2017ESE}. Second, NNs are theoretically universal approximators to fit any continues function~\cite{Hornik1991Approximation}. Previous works mimic the functions of approximable code regions~\cite{esmaeilzadeh2012neural} using NNs. Third, NNs expose enormous parallelism for performance/energy benefits, in favor of the booming development of hardware accelerators~\cite{Liu-isca2016}. 
The same accelerator can approximate various functions by merely changing the NN topologies and weights. Above advantages make NNs a principal candidate in approximate computing, namely~\emph{approximators}.

Quality control is essential in approximate computing to identify those safe-to-approximate work loads~\cite{khudia2015rumba}. On one hand, various techniques employ high-level, application specific light-weight checks(LWCs) to estimate the quality of the approximation output~\cite{grigo2015lwc}. These works then advocate a rollback or online tuning mechanism to guarantee the computation reliability. On the other hand, a proactive strategy is to predict whether the load is safe-to-approximate or not using statistical and learning models~\cite{sampson2014expressing, khudia2015rumba,Wang-islped16,sui2016proactive,Rahimi2016An}. 


Approximate computing framework that deploys two neural networks can further optimize the approximation quality and invocation~\cite{mahajan2016towards, Xu2017On}. In these works, one neural network is trained as a \emph{classifier}, to differentiate safe-to-approximate input from others, while the other as an \emph{approximator}. It is challenging to train two neural networks because of the correlation between them, but we advocate this approximate computing framework in this paper due to the superiority of neural networks.


Existing neural approximate computing with quality control strives to optimize two objectives simultaneously: minimizing the approximation error and maximizing the invocation of approximator~(the probability of invoking the approximator to gain energy efficiency). In this work, we argue that these two goals may contradict each other. Based on the principle of approximate computing, it is not necessary to further reduce the approximation error when the error is lower than the error bound. Instead, we should maximize the invocation of approximator. Because more input invokes the accelerator---ASIC~\cite{esmaeilzadeh2012neural}, FPGA~\cite{Moreau2015SNNAP} or NVM~\cite{Li2015RRAM,chi2016prime} etc.---rather than CPU, more energy efficiency we gain. Therefore, we advocate the invocation-driven design for approximate computing. In this paper, we propose a novel approximate computing architecture \emph{to maximize the invocation of approximate accelerators. The key idea is to orchestrate multiple approximators which fit more input data.} The contribution of this work is summarized as below:
\begin{itemize} 
\item We investigate two compound structures of approximate computing: i)  Multiple cascaded classifiers and approximators~(MCCA) architecture that naturally extends from the existing methods
; ii) Multiclass-Classifier and Multiple Approximators architectures~(MCMA) that processes much faster than MCCA. 

\item We investigate two co-training methods, i.e. complementary and competitive training, for multiple approximators in the compound structure. Each approximator can fit a distribution of the input data---a partition in input space---and the whole compound structure can theoretically fit all samples in the input space thus maximize the invocation of approximate accelerators. 

\item We implement the compound approximate computing structures in the existing neural processing unit(NPU) design. Based on the investigation, the proposed approximate computing architecture can achieve superior speedup without hardware or performance overhead.
\end{itemize}

The rest of the paper is organized as follows: section~\ref{sect:preliminary} shows related works and motivation. Proposed approximate computing architectures are described in section~\ref{sect:approach}. Section~\ref{sect:exp} shows the experiments, and section~\ref{sect:conclusion} concludes this paper. 

\section{Related works and Motivation}\label{sect:preliminary}

\subsection{Model based quality control for Approximate Computing}\label{ssect:related_quality}
Various approximate computing methods use statistical or learning models to predict the approximation errors. Rumba et al.~\cite{khudia2015rumba} propose decision trees and linear models for predicting the error in the runtime. Wang et al.~\cite{Wang-islped16} propose a reinforcement learning method to determine when and how to rollback for occasional large errors with a minimum cost. However, due to the bias of input feature space, some approximate outputs cannot satisfy the quality requirement. These works give up screening the input data on the accelerator call site, resulting in unnecessary rollbacks.

Proactive quality control mechanisms use predictors to determine the input shifted to the accelerator and invoke the accelerator. In~\cite{sui2016proactive}, a Bayesian network learns the cost and error models of an optimization problem offline and determines the quality control knob by solving this optimization problem. Rahimi et al.~\cite{Rahimi2016An} selectively reduce and dynamically tune the precision, subjecting to a statistical quality knob; consequently, an approximator within the fixed area budget can accommodate more parallel approximate kernels. These works strive to maximize the invocation of a single approximation scheme. \emph{The upper limit of a given approximation scheme, e.g., an 8-bit precision approximate-multiplier, undoubtedly circumscribes the potential of safe-to-approximate input in the input space, preventing us from exploring more energy efficiency.}

\subsection{Approximate Computing with two neural networks}\label{ssect:related_twoNNs}
Mahajan et al.~\cite{mahajan2016towards} present the first approximate computing framework that employs two neural networks as the classifier and the approximator, respectively. They make the best effort to train an approximator using the input and output data of the target function for approximation. Then, they test the approximator using the same input data and generate the approximation output. Comparing the approximation error with the error bound can tell us whether the input data is safe-to-approximate or not, which serves as labels to train the classifier.  Their results show the NN-based predictor is better than the table-based predictor regarding prediction accuracy. However, this work ignores the correlation between the two neural networks. The capability of the classifier is constrained by the approximator because the approximator produces the ``label'' in the training sample of the classifier.   


Consequently, Xu et al.~\cite{Xu2017On} propose an iterative training method for the classifier and approximator. The fundamental idea is to repeat the training process in~\cite{mahajan2016towards}, retrain the approximator using only the safe-to-approximate input identified by the classifier, and then retrain the classifier again using the new ``label'' produced by the approximator. As the iterative training goes on, the approximation error of the approximator and the prediction error of the classifier keep decreasing. However, this method is weak in enhancing the invocation of the approximator because the amount of safe-to-approximate data shrinks after several iterations. 

\subsection{Motivation}

\begin{figure}[t!]
    \centering
    \includegraphics[width=1\linewidth]{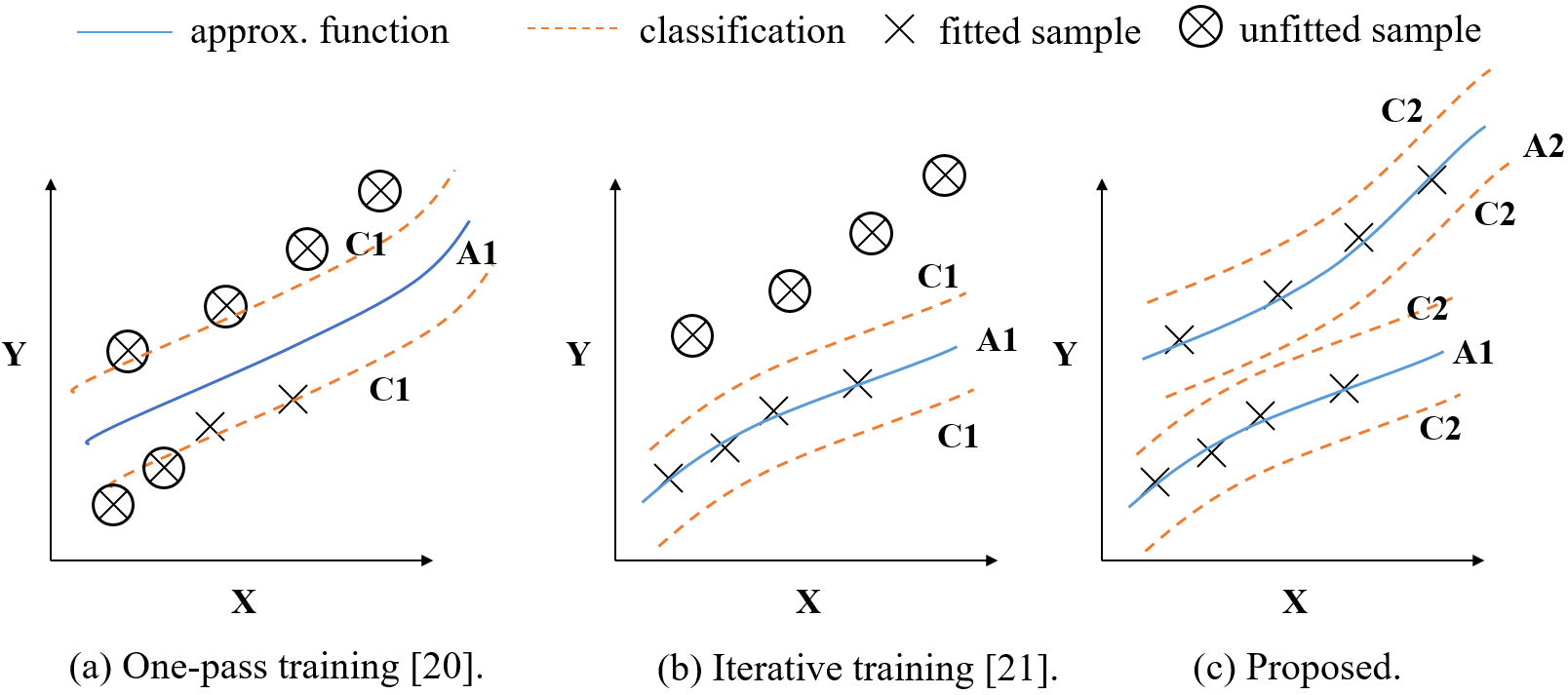}\vspace{-10pt}
    \caption{A conceptual view of fitting the data samples with different neural approximate computing architectures.}\label{fig:comp-training}
    \vspace{-15pt}
\end{figure}

Without the loss of generality, we use a simple but motivational example to point out the remaining problem of the existing neural approximate computing framework involving two neural networks.  In~\cite{mahajan2016towards}, the approximator~(\emph{A1}) and the classifier~(\emph{C1}) are trained separately in one pass, as shown in Fig.~\ref{fig:comp-training}(a). The optimization targets of \emph{A1} and \emph{C1} may be mismatched. The \emph{A1} strives to fit ``all'' possible input samples, which makes it challenging for \emph{C1} to differentiate the safe-to-approximate data from others. By the iterative retraining of \emph{A1} using the safe-to-approximate input samples recognized by \emph{C1}, and vice versa, as shown in Fig.~\ref{fig:comp-training}(b), \emph{A1} and \emph{C1} can coordinate their optimization objectives~\cite{Xu2017On}. \emph{A1} evolves to provide more accurate approximation output, while \emph{C1} becomes more robust to discriminate those safe-to-approximate data from others. However, the approximator after sufficient training may overfit one cluster/distribution of input sample. Such bias also pushes away other input samples from the approximation function. The input sample pushed out of the ``error bound'' in the view of the classifier results in a degraded invocation of the approximator. 

Motivated by this, the primary target of this work is to maximize the invocation of the approximator~(salvage the abandoned safe-to-approximate data), by initiating multiple approximators. As shown in Fig.~\ref{fig:comp-training}(c), each approximator~(e.g., \emph{A1} and \emph{A2}) concentrates on fitting a reasonable amount of data samples that are easily recognized by the classifiers~(e.g., \emph{C1}, \emph{C2}). The compound structure can cover much more data in the input space that maximizes the overall invocation. The rest of this paper focuses on designing a compound structure of neural networks and their training methods. 


\begin{figure}[b!]
    \centerline{
		\subfigure[Training with \emph{category C}.]{
            \includegraphics[width=0.5\linewidth]{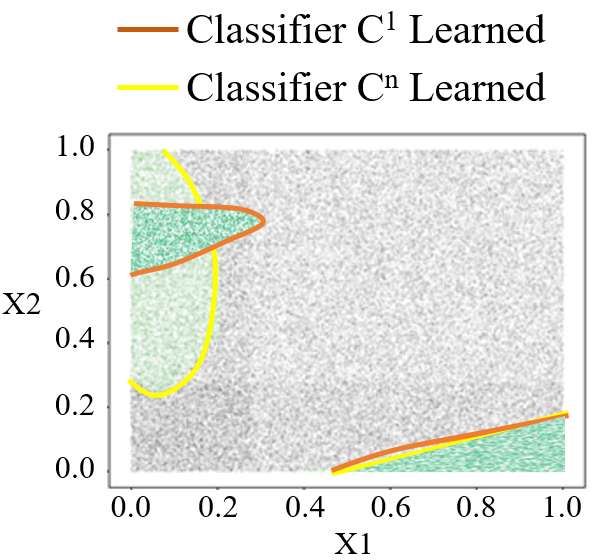}
		}
		\subfigure[Training with  \emph{category A}.]{
            \includegraphics[width=0.5\linewidth]{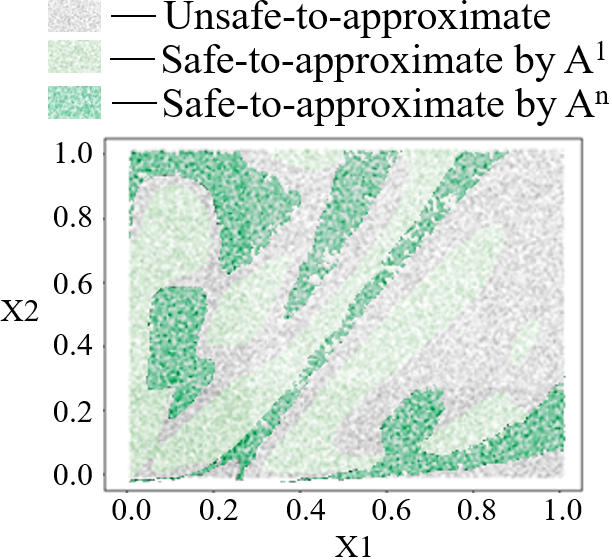}
		}
	}\vspace{-10pt}
    \caption{Distribution of training data for approximator in iterative training. $C^1$ and $C^n$ mean classifier in the first and the $n_{th}$ iteration. The same for $A^1$ and $A^n$.}
    \label{fig:Category_A_C}
\end{figure}


\section{Proposed Approximate Computing Architecture}\label{sect:approach}
In this section, we analyze and interpret different data selection strategies based on iterative training method~\cite{Xu2017On}.  The observation inspires us to develop 
two compound structures of neural networks for our approximate computing architecture. At last, we deploy our architectures in hardware design based on a typical Neural Processing Unit design~\cite{esmaeilzadeh2012neural}. 



\subsection{Clustering of safe-to-approximate input sample}\label{sect:moti}

It is challenging to optimize the invocation of the approximator because there is no place to put the invocation into the loss functions of both neural-networks. Two neural networks have two viewpoints upon the input samples. We can select training data predicted as safe-to-approximate by the classifier in the current iteration, denoted as category~\emph{C}, or choose samples whose approximation error--resulting from the approximator--is within the error bound, denoted as category~\emph{A}. Such difference is due to the mismatch between the viewpoints of the two neural networks. The previous work~\cite{Xu2017On} simply chooses the safe-to-approximate samples, on which the two neural networks agree~(denoted as ``AC''). 
Alternatively, we discover the clustering effect of the safe-to-approximate samples that can help us increase the invocation of the approximator. 

The approach to such discovery is described as follows. Given a typical RMS application, we apply iterative training multiple times, and plot the input samples: safe-to-approximate samples are in green color. We further track the change of the plots in the final iteration and color the safe-to-approximate input samples in light green. As depicted in Fig.~\ref{fig:Category_A_C}~(a) and (b), the safe-to-approximate input samples are clustered when selecting the training data using \emph{C} in the iterative training process; the sphere of these clusters gradually expands in the later iterations of training. When choosing \emph{A} in each iteration of training, on the contrary, the safe-to-approximate input samples are scattered into the input space as small pieces. It is much easier for the classifier to discriminate the clustered safe-to-approximate samples. It is also easier for approximator to fit the clustered safe-to-approximate samples. 

For those samples out of the clusters, we can remedy those abandoned samples using additional approximators. Therefore, we orchestrate multiple approximators in the approximate computing architecture and develop the training methods to incline each approximator to a different cluster of safe-to-approximate samples, as described in the next section. 

\begin{figure*}[t!]
    \centerline{
        \subfigure[Training.]{
            \includegraphics[width=0.6\textwidth]{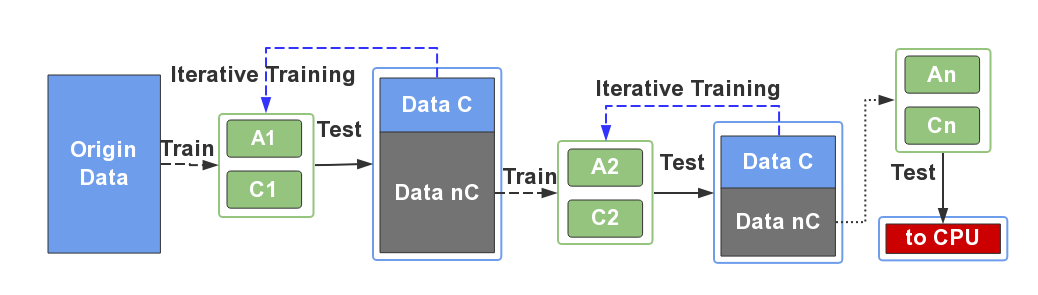}\vspace{-10pt}
        }
        \subfigure[Execution.]{
            \includegraphics[width=0.4\textwidth]{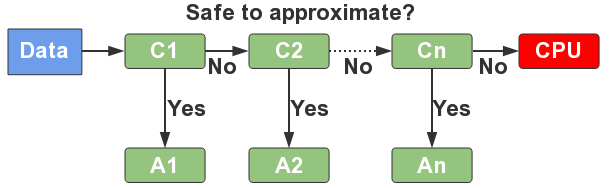}\vspace{-10pt}
        }
    }
    \caption{Proposed MCCA architecture and its training process.}
    \label{fig:mcca-training}
    \vspace{-10pt}
\end{figure*}
\subsection{Multiple Cascaded Classifiers and Approximators~(MCCA)}\vspace{-2pt}


The key idea to increase the invocation of approximator is to consecutively employ additional approximator for the remedy of the input samples abandoned in previous iterations. Meanwhile, a question emerges: how many approximators are enough to cover the majority of the input space?


To answer this question, we propose the Multiple Cascaded Classifiers and Approximators (MCCA) structure by cascading multiple pairs of the approximator-classifier structure proposed in the iterative training method. Fig.~\ref{fig:mcca-training}(a) illustrates an example MCCA structure and the training process. MCCA structure is composed of cascaded pairs of classifier and approximator, each of which sequentially divides input data space into safe-to-approximate and unsafe-to-approximate clusters. In the beginning, the original input samples are used to train the first pair of the classifier-approximator structure, denoted as $A_1$ and $C_1$. The training process is similar to the iterative training~\cite{Xu2017On}, except that we select the training samples using category \emph{C} to train the $C_1$-$A_1$ pair \emph{in the second iteration.} After the above training process converges, we feed the remaining input samples not yet to be recognized by $C_1$~(Data nC) to the second pair $C_2$-$A_2$. Such process continues until a specific pair of $C_n$ and $A_n$ cannot converge. The remaining unsafe-to-approximation data should enter CPU for precise computation.




At run-time, the approximate computing architecture with MCCA structure also consumes the input loads in a cascaded manner. The execution process is shown in Fig.~\ref{fig:mcca-training}(b). $A_1$ approximates the input data based on the prediction of $C_1$. If $C_1$ disapproves, the input data are sent to the next pair $C_2$-$A_2$. The input data are finally executed in CPU if rejected by all the classifiers. Obviously, MCCA is too time consuming that may offset the speedup resulting from the approximate accelerator. 
We solve this problem by proposing a more efficient structure in the next section. 


\subsection{Multiclass-classifier and Multiple Approximators~(MCMA)}

\begin{figure}[b!]
    \centerline{
         \subfigure[Execution.]{
            \includegraphics[width=0.25\linewidth]{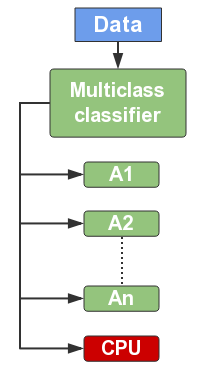}\vspace{-10pt}
            \label{fig:MCMA_exeution}
        }
         \subfigure[Training process.]{
            \includegraphics[width=0.75\linewidth]{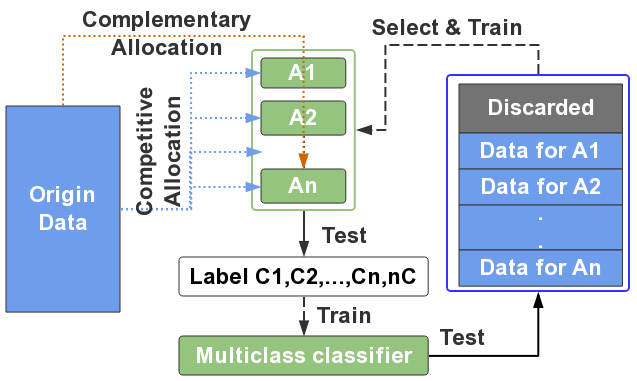}\vspace{-10pt}
            \label{fig:MCMA_train}
        }
    }
    \caption{Proposed MCMA architecture and its training process.}
    \label{fig:MCMA}
    \vspace{-5pt}
\end{figure}

We propose a ``parallel'' compound structure with a multiclass-classifier and multiple approximators~(denoted as MCMA). Given the input samples at the runtime, as shown in Fig.~\ref{fig:MCMA_exeution}, the multiclass-classifier predicts which approximator can generate a safe-to-approximate result. The approximator with the highest confidence in the prediction consumes the input sample. A critical question is how to train such a paralled structure.

We propose two data allocation mechanisms to train the MCMA structure. For \emph{Complementary} mechanism, input samples are fed into all the approximators from $A_1$ to $A_n$ serially. The latter approximator is trained using the input samples that previous approximators fail to approximate. After the initialization, the training process of MCMA structure adopts the iterative training method.  In each iteration afterward, we generate the label for training the multiclass-classifier using the \emph{complementary} mechanism: $A_1$ tests all the input samples and produces the label~$C_1$ for any input sample that $A_1$ can safely approximate. Otherwise, no label is assigned to that input sample.  Subsequently, $A_2$ tests the remaining input samples without any label and produces label~$C_2$ for the sample that is safe-to-approximate for $A_2$. This procedure continues until all the approximators have finished the testing. The remaining input samples without any label are labeled as~$nC$. We then train the multiclass-classifier using all the input samples with labels. In the next iteration, we test the derived classifier using all the input samples. The resulting prediction, e.g., $Ci$, with the highest confidence indicates the most suitable approximator, e.g., $Ai$, to approximate the input sample. The multiclass-classifier distributes each input samples to its most suitable approximate and eventually partitions the input space into $n+1$ territories. Each of the \emph{n} approximators takes the samples in its territory for the next iteration of training. The basic idea of the complementary allocation mechanism is similar to \textbf{AdaBoost} algorithm~\cite{Freund1997Adaboost}. However, the execution process of the MCMA structure is more efficient than that of an AdaBoost algorithm: The MCMA structure derives the final approximation from one approximator, while an AdaBoost algorithm derives the result by voting from all the approximators. 

For the \emph{Competitive} mechanism, in the first iteration, we assign all the input samples to all the approximators in parallel. All the approximators compete with each other to fit as many input samples as possible. Each approximator may bias different distribution of the input samples due to the randomness in the training samples. 

Furthermore, we can vary the hyper-parameters in the neural networks such that each approximator can reach different local minima. After the initialization, we competitively train the multiclass-classifier: each approximator tests each input sample and generates the approximation error. We generate the label for this sample according to the approximator deriving the lowest approximation error and use this label to train the multiclass-classifier. Then, we test the multiclass-classifier and assign the samples to the approximator that the classifier trusts the most. The same procedure continues iteratively. 

Note that multiple approximators may have the similar confidence to approximate the same input sample. Their territories may thus overlap to each other. After some iterations, the bias of each approximator is reinforced, and the multiclass-classifier can easily distinguish the territories of different approximators. 

\subsection{Hardware design of MCMA structure}

\begin{figure*}[htb]
    \centering
    \subfigure[The NPU structure and the data flow of MCMA.]{\includegraphics[width=0.416\linewidth]{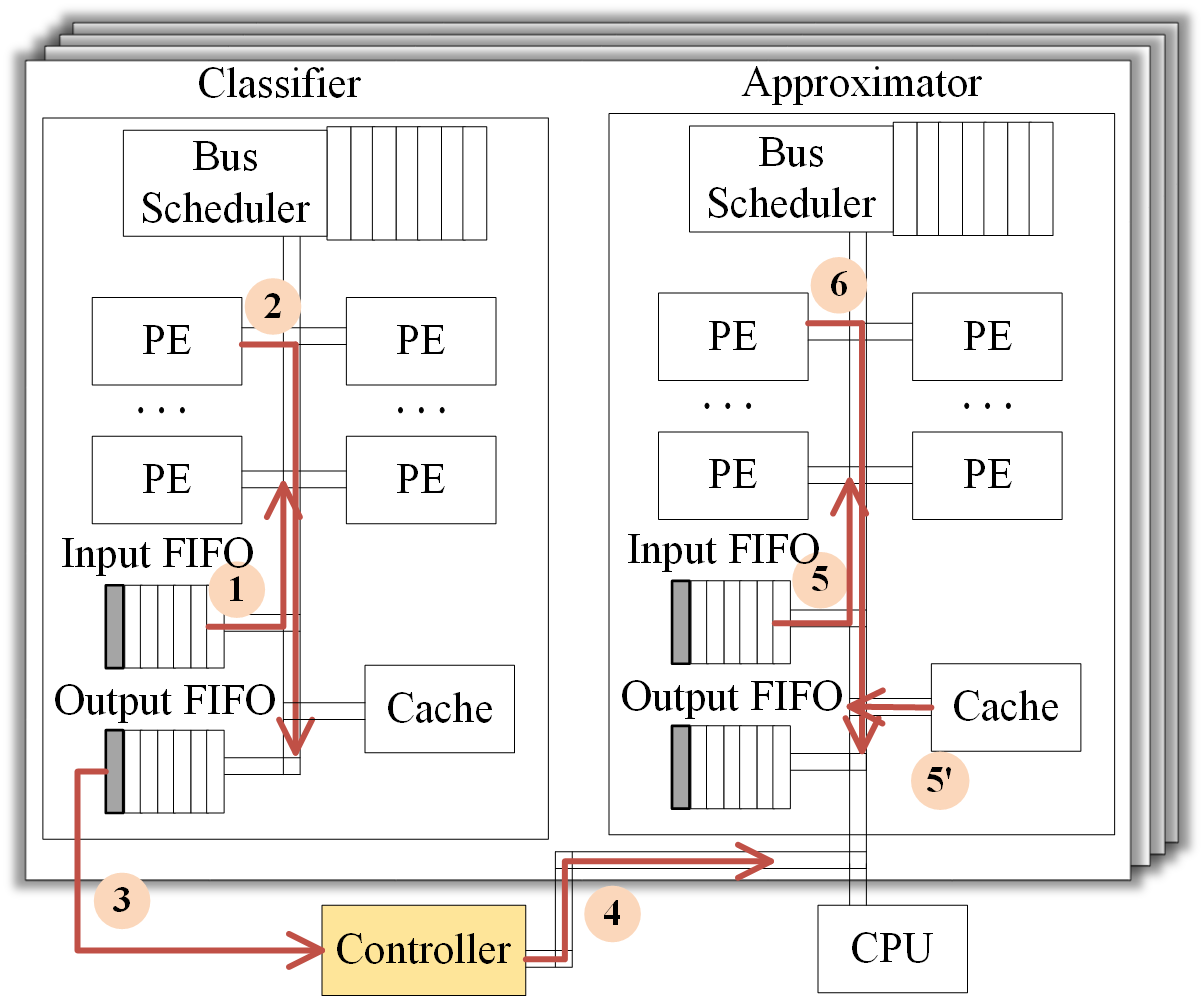}
    \label{fig:NPU_Arch}
    }
    \subfigure[Structure of a PE.]{\includegraphics[width=0.336\linewidth]{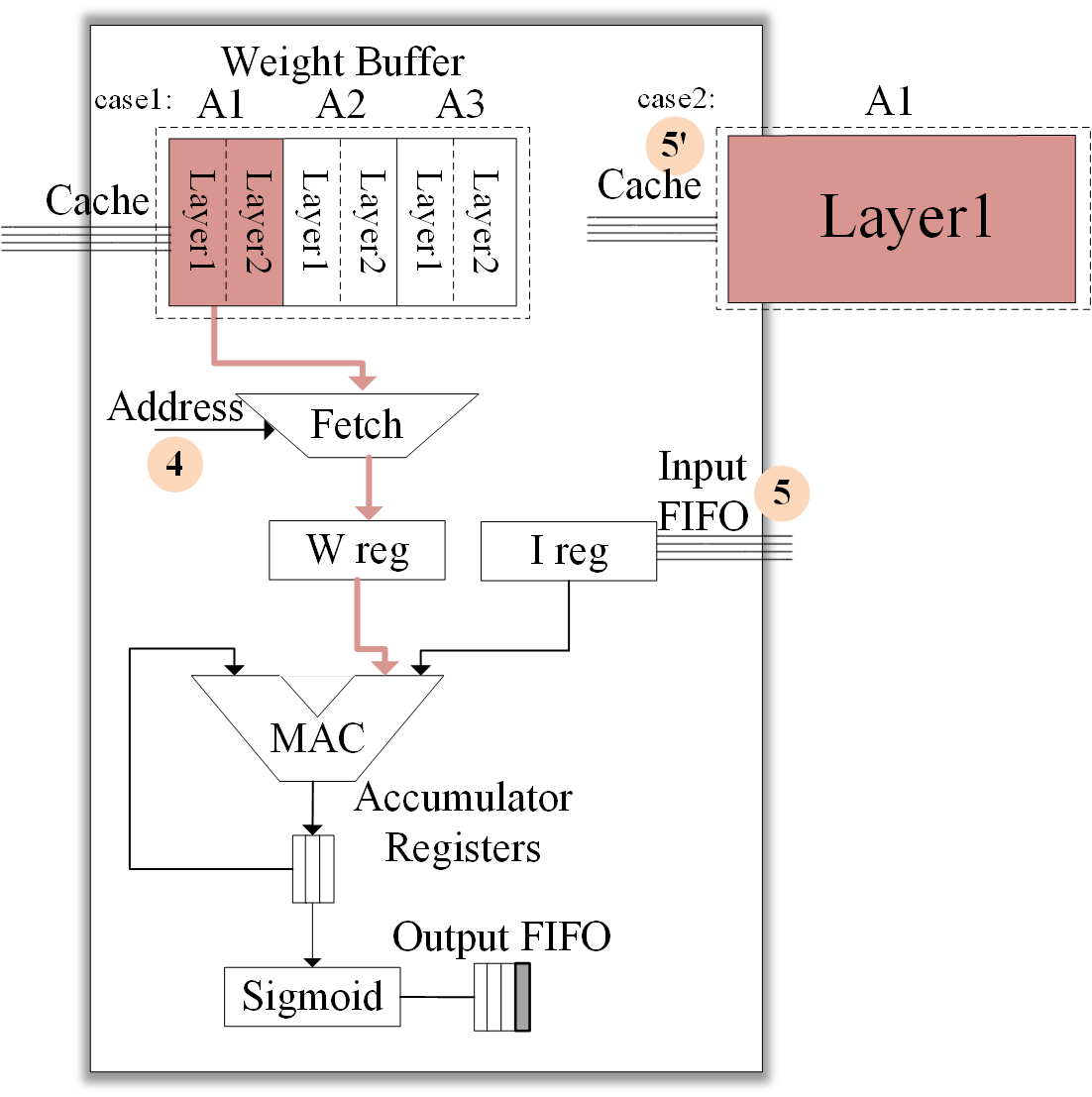}
    \label{fig:NPU_Arch_PE}
    }
    \caption{Proposed NPU Architecture.}
    \label{fig:NPU_Architecture}
    \vspace{-15pt}
\end{figure*}

The \emph{MCMA} architecture consists of more than one approximators. Thus, we propose an NPU design that provides instant switch among different approximators and enlarges the degree of parallelism. As shown in Fig.~\ref{fig:NPU_Arch}, the NPU consists of identical tiles of computing resource based on~\cite{esmaeilzadeh2012neural}. Each tile includes multiple Processing Elements~(PEs), Input/output FIFOs, and a Cache, connected by an internal bus. PE calculates the output of one neuron at a time in the inference task of the neural network. The bus scheduler schedules the data from input FIFO to PEs and from PEs to output FIFO through the data bus. It also schedules the weights of the neural network from the cache to each PE. 

The structure of the PE, as shown in Fig.~\ref{fig:NPU_Arch_PE}, includes a weight buffer, a fetch unit that reads the weights from the specific addresses of the weight buffer and sends the weights to W register. The I register stores the input sample transferred from Input FIFO. When both values arrive in the registers, a Multiply Add Accumulator unit carries out the arithmetic computation. The result is activated by the activation unit~(e.g., Sigmoid) and sent to Output Buffer. We can dynamically allocate the classifier and approximators to the tiles according to the scale of resource and neural networks. Considering the multiple approximators, we deploy a controller which receives the prediction result from the classifier and sends the control signal to the approximator.

Fig.~\ref{fig:NPU_Architecture} shows the data flow. The weights of the classifier are loaded in the initialization stage. In stage 1, data are transferred from the input FIFO to each PE of the classifier. In stage 2, each PE calculates the output of one neuron and put the output in the output FIFO. In stage 3, the result is sent to the controller from the output FIFO. In stage 4, the controller invokes the approximator if the data are safe-to-approximate, otherwise invokes the CPU. Then in stage 5, the input data are transferred to PEs in the approximator for computation. In stage 6, the results are sent back to the output FIFO. 

Our architecture needs to support Weight switch among different approximators, as shown in Fig.~\ref{fig:NPU_Arch_PE}. Three scenarios may occur according to the size of the neural networks. Case 1: the weight buffer can store all weights of the approximators. No weights need to be loaded from the cache. 
Case 2: the weight buffer is not large enough to store the weights of one approximator. In this case, either the \emph{MCMA} method or any other NPU needs to load the weights layer by layer. So there is no extra overhead compared with previous methods. Case 3: the weight buffer is large enough for storing one approximator but not enough for storing all approximators. In this case, when the prediction of the $i_{th}$ sample is different from the $i-1_{th}$ sample, the approximator loads the weights from cache to the weight buffer.

In summary, the MCMA architecture can adapt to any existing NPU design such as~\cite{esmaeilzadeh2012neural} by adding a simple controller for switching the approximations. The data movement between the NPU and other components, e.g., CPU or DMA, remains the same as the original NPU design.

\section{Experiments and Results}\label{sect:exp}
\subsection{Experimental setup}

In the experiment, we compare the proposed MCCA and MCMA architectures, to the conventional classifier-approximator architecture~(denoted as \emph{Iterative}~\cite{Xu2017On} and \emph{one-pass}~\cite{mahajan2016towards}). 
For all classifiers and approximators, we use multilayer perceptrons with backpropagation algorithm. \emph{MCCA}, \emph{MCMA} and \emph{Iterative} are trained with five iterations. And \emph{one-pass} with one iteration. In the training, we use RMSprop optimizer and set epoch as 1500.


\begin{figure*}[t]
    \centering
    \includegraphics[width=1\linewidth]{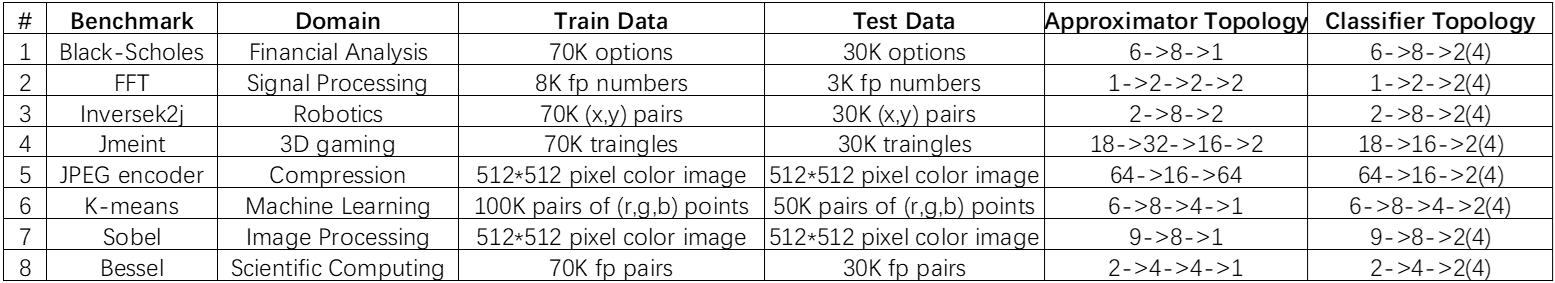}
    \vspace{-15pt}
    \caption{Benchmark description}\vspace{-15pt}
    \label{fig:benchmark_discription}\vspace{-5pt}
\end{figure*}

We select eight benchmark applications from~\cite{esmaeilzadeh2012neural} and GNU GSL scientific computing library as shown in Fig~\ref{fig:benchmark_discription}. The first seven benchmarks are identical from the previous paper~\cite{esmaeilzadeh2012neural}. These benchmarks are for comparison among different methods. The Bessel benchmark has two-dimensional input, thus we use images to show the data distribution. In the table, the topology of the neural network is described by the number of neurons in each layer. For example, $6\rightarrow8\rightarrow1$ means there are 6 neurons in the input layer, one hidden layer with 8 neurons and an output layer with one neuron. In every benchamrk, the approximators and classifiers in all methods are of the same topology, except the last layer of the classifiers in \emph{MCMA} method, due to the multi-classification task. The network topologies are selected by balancing between the performance and the structure size. 

We use invocation of classifiers and root-mean-square error(RMSE) of the data approximated by the approximator(simply we call it \emph{error}) to measure our models. The error bound means the quality requirements for the output. The lower the error bound, the higher the quality requirement. 
We vary the error bounds and show the results considering the various quality requirements among different applications. We also visualize the statistics of the data distribution.

\subsection{Results and analysis}

Compared with previous methods, our architecture has i) higher invocation of the approximators while keeping the \emph{error} under the threshold; ii) higher speedup and energy efficiency.

\begin{figure}[h!]
    \centering
    \subfigure[Comparisons on the invocation across different benchmark functions.]{
        \includegraphics[width=1\linewidth]{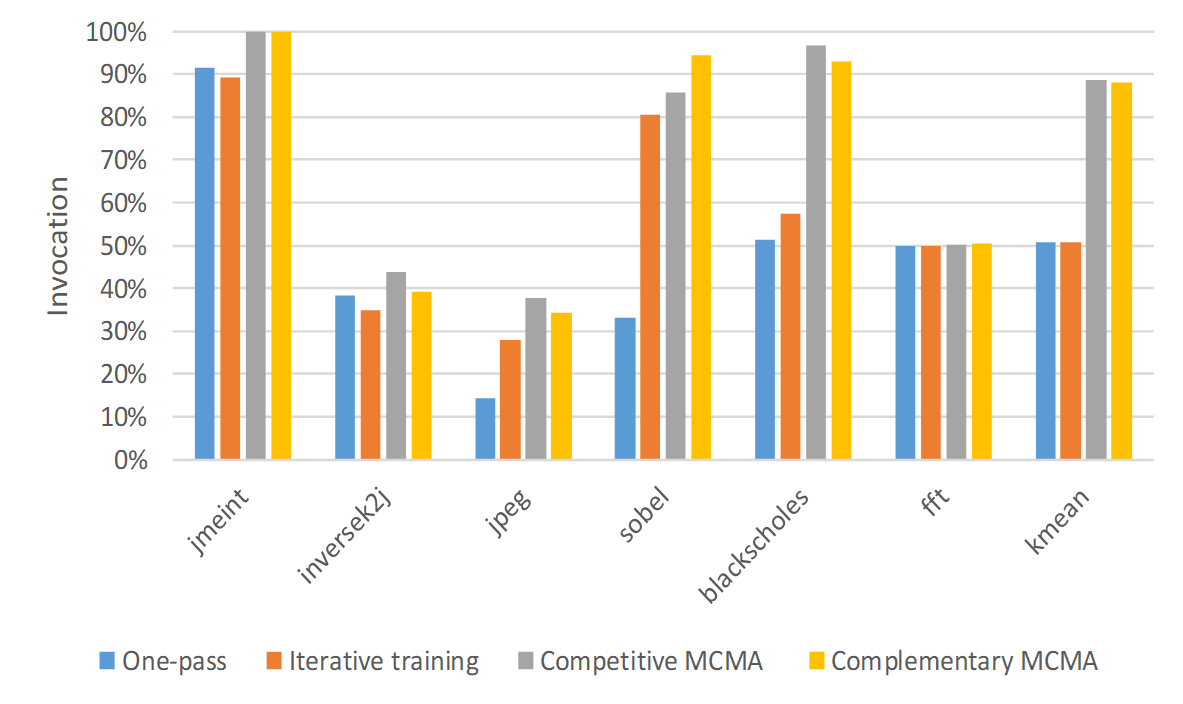}
        \label{fig:invocation_graph}\vspace{-10pt}
    }
    \centering
    \subfigure[Comparisons on the approximation error normalized to the error bound.]{
        \includegraphics[width=1\linewidth]{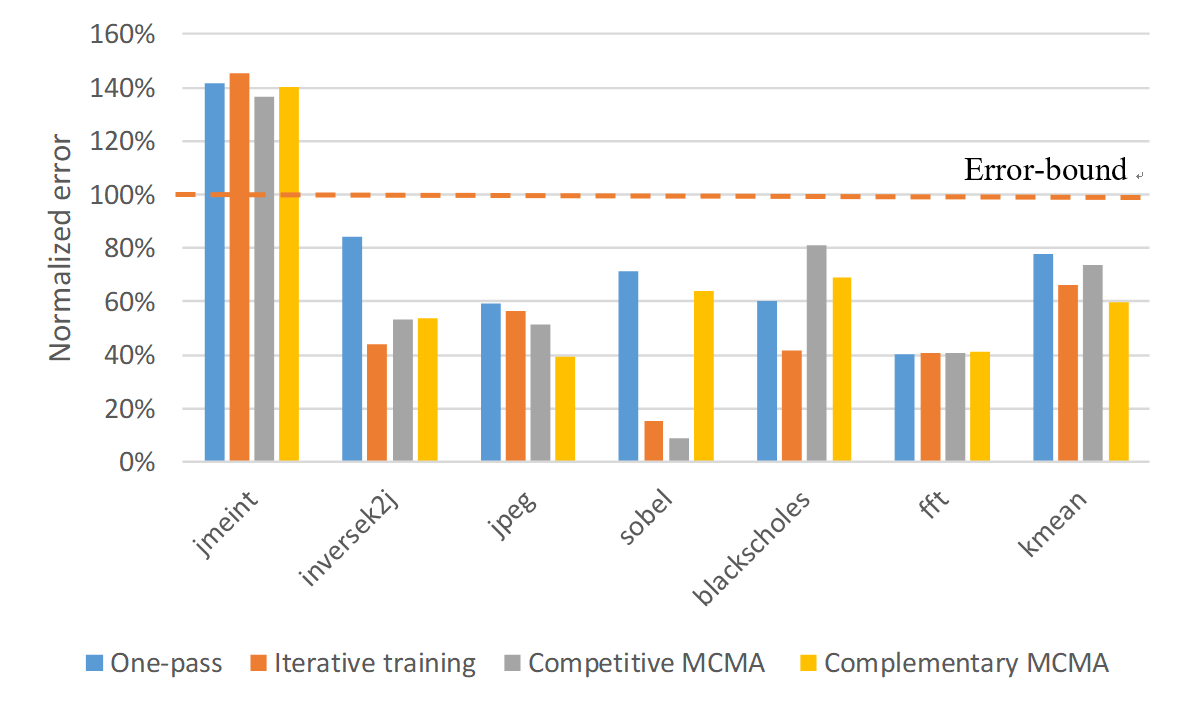}
        \label{fig:error_graph}\vspace{-10pt}
    }
    \centering
    \subfigure[Comparisons on the invocation varying the error bound in Black-Sholes.]{
        \includegraphics[width=1\linewidth]{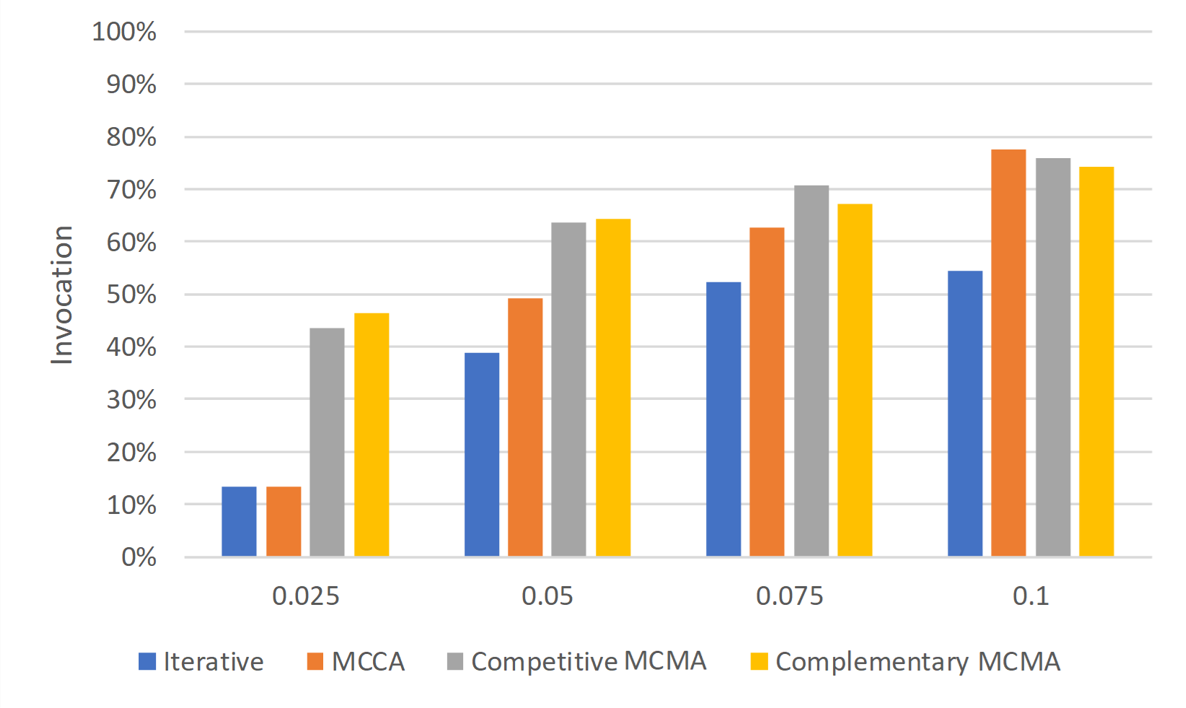}
        \label{fig:exp_errorbound}\vspace{-15pt}
    }
    \caption{Results of invocation and error.}\vspace{-10pt}
\end{figure}

Fig.~\ref{fig:invocation_graph} and Fig.~\ref{fig:error_graph} show the ``invocation'' and ``approximating error'' in different benchmarks among \emph{one-pass}, \emph{iterative} and two types of \emph{MCMA} architectures. We normalized the approximation error with respect to the error bound.
In average, the \emph{MCMA} methods have greater invocation than the \emph{one-pass} method by $27\%$ and less ``approximation error'' by $10\%$. In kmeans and Black-Scholes, our methods outperform the previous methods in invocation by $40\%$(from $50\%$ to $90\%$) while the ``approximation error'' remains unchanged. In most benchmarks, the approximation error is below the error bound which means the quality control is good except the jmeint benchmark. The FFT bench is regarded as "not suitable for approximation" and thus all the methods show no difference.

Fig.~\ref{fig:exp_errorbound} shows the detailed results on the Black-Sholes benchmark by varying error bound. Invocation of all the methods increases as the error bound arises. 
When the user requires a tighter error bound, compared with other methods, the drop of invocation of our proposed architecture is the smallest. In other words, the proposed architecture is more desired for those approximate critical applications. 
In most cases, the \emph{MCMA} architecture has a higher invocation of the approximator than the \emph{MCCA}. Also since the \emph{MCCA} architecture is cascaded which is not time and energy efficient, the rest of the experiment will focus on the \emph{MCMA} architecture.

\begin{figure}[htb]
    \centering
    \subfigure[Comparison on speedup normalized to one-pass method.]{
        \includegraphics[width=1\linewidth]{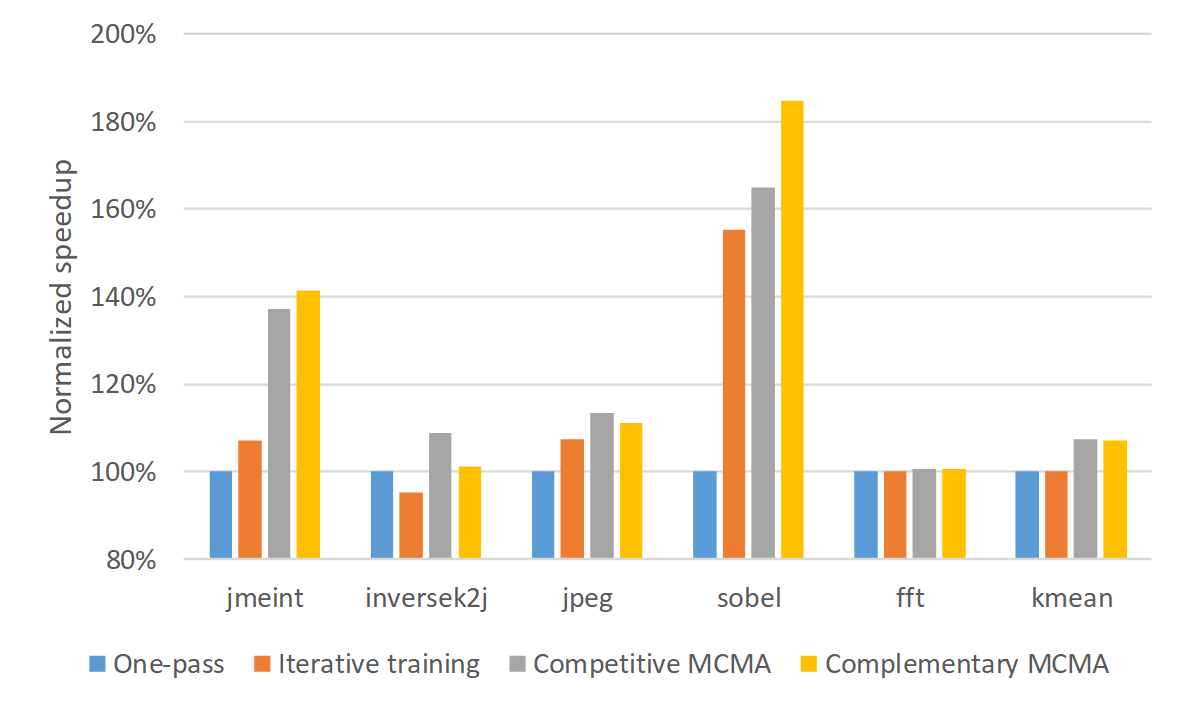}
        \label{fig:speedup_graph}\vspace{-10pt}
    }
    \centering
    \subfigure[Comparisons on energy reduction normalized to one-pass method.]{
        \includegraphics[width=1\linewidth]{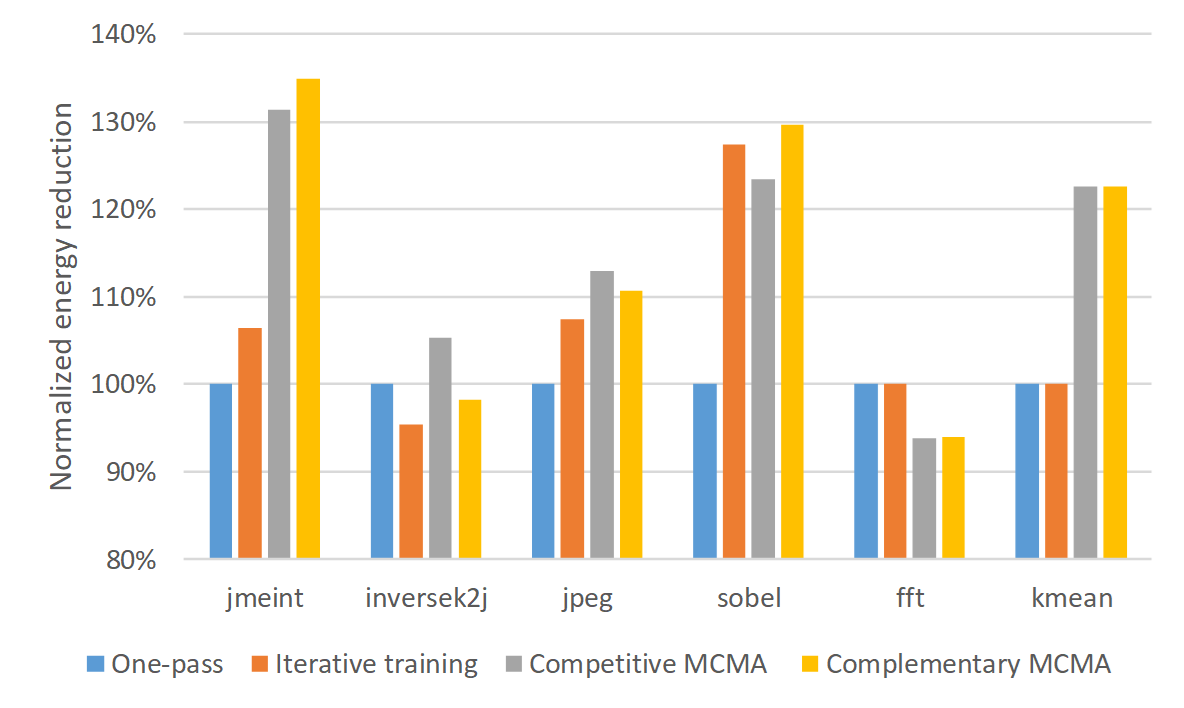}
        \label{fig:energy_graph}\vspace{-10pt}
    }
    \caption{Comparisons on speedup and energy reduction.}\vspace{-10pt}
    \label{fig:speedup_energy}
\end{figure}

Fig.~\ref{fig:speedup_energy} shows the ``speedup'' and ``energy reduction'' in different benchmarks corresponding to the invocation and error in Fig.~\ref{fig:invocation_graph} and Fig.~\ref{fig:error_graph}. The results are normalized respect to the \emph{one-pass} method. Due to the space limit, we estimate the performance of MCMA by scaling the performance of NPU in~\cite{esmaeilzadeh2012neural} based on the invocation of NPU. This is a valid estimation as the proposed MCMA architecture is merely the same as original NPU design.
The average speedup is about $1.23\times$ and the energy reduction is about $1.15\times$ in both \emph{competitive MCMA} and \emph{complementary MCMA}. The computation gap between CPU and NPU is large, so the speedup is largely determined by the invocation for those computation-bound applications. For those communication-bound applications, neural approximate computing may not be suitable. In the sobel benchmark, the invocation of \emph{MCMA} method is much larger than the \emph{one-pass} method, leading to significant speedup and energy reduction.

\begin{figure}[htb]
    \centering
    \includegraphics[width=1\linewidth]{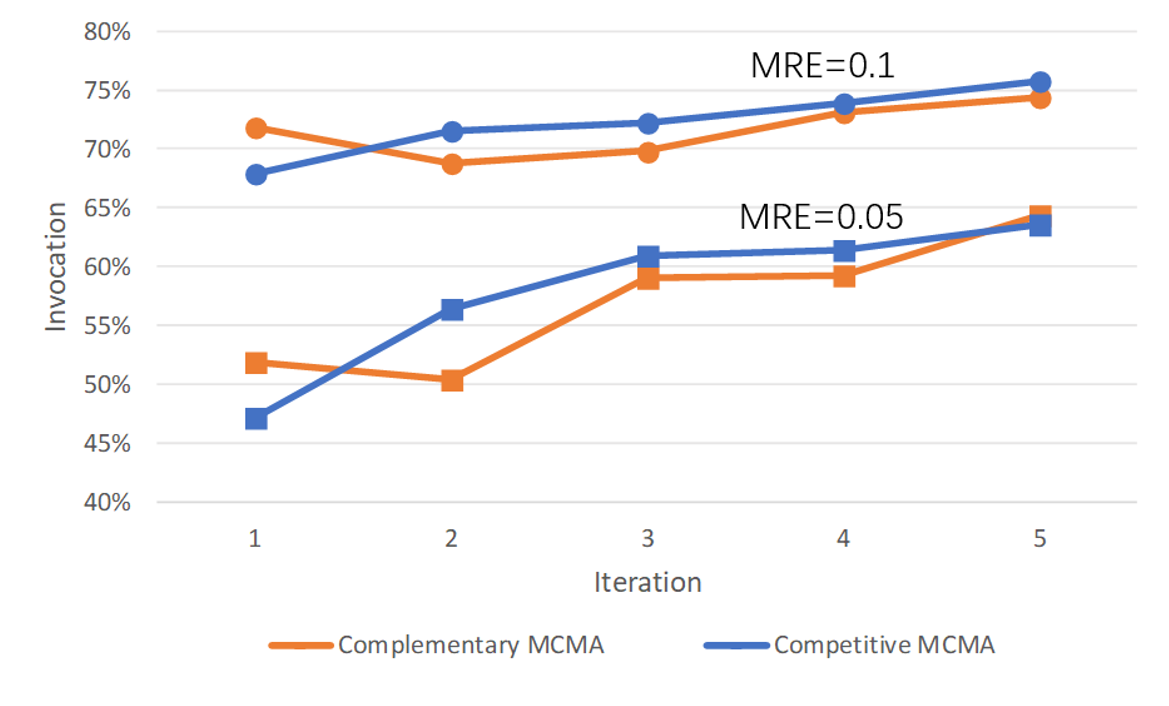}\vspace{-10pt}
    \caption{Comparisons on Invocation between \emph{Complementary} and \emph{Competitive} allocation schemes in MCMA.}
    \label{fig:itr}\vspace{-5pt}
\end{figure}
\begin{figure}[htb]
    \centering
    \subfigure[Distribution of data samples of three approximators.]{
    \includegraphics[width=0.7\linewidth]{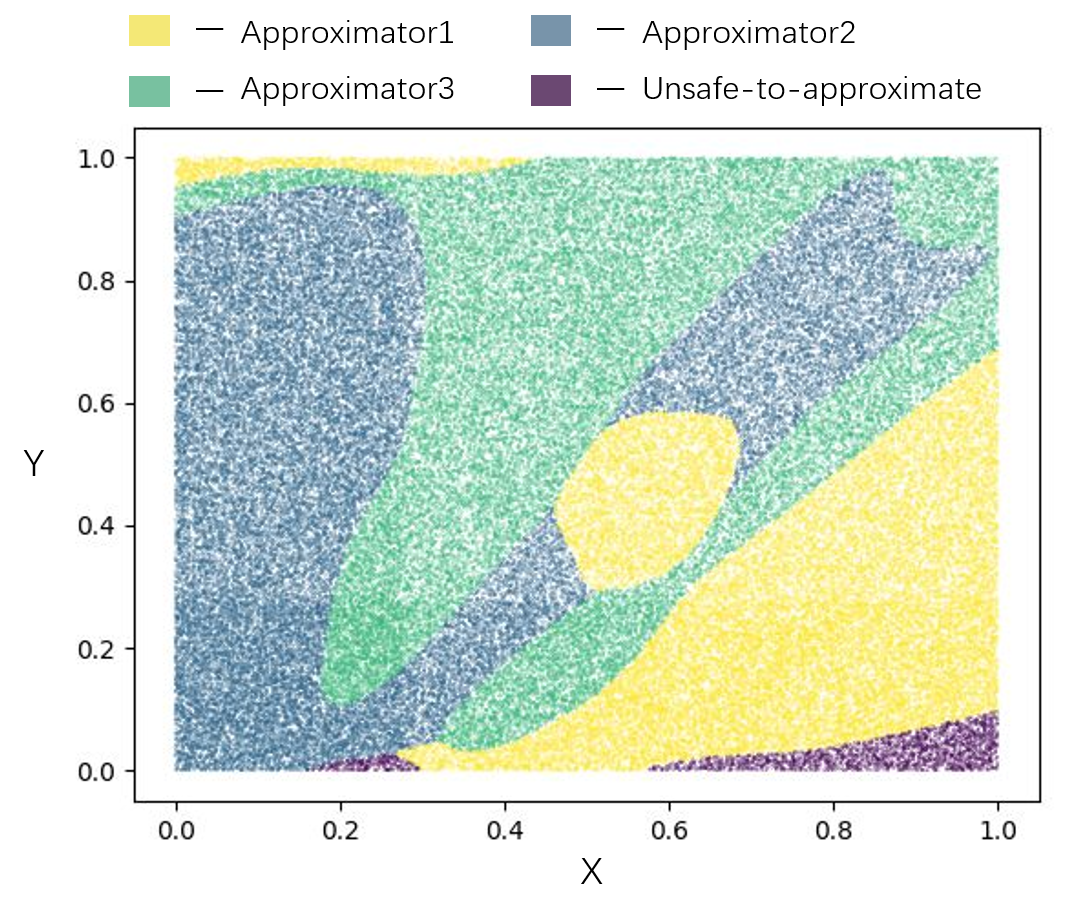}
    }
    \subfigure[Relative error derived by the three approximators.]{
       \includegraphics[width=0.7\linewidth]{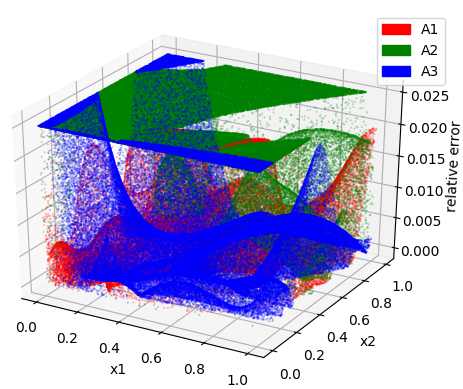}
    }
    \caption{Data distribution of the Bessel bench using the MCMA architecture.}\label{fig:3dapprx}\vspace{-15pt}
\end{figure}

\begin{figure*}[thb]
    \centering
    \subfigure[one-pass method]{\includegraphics[width=0.3\linewidth]{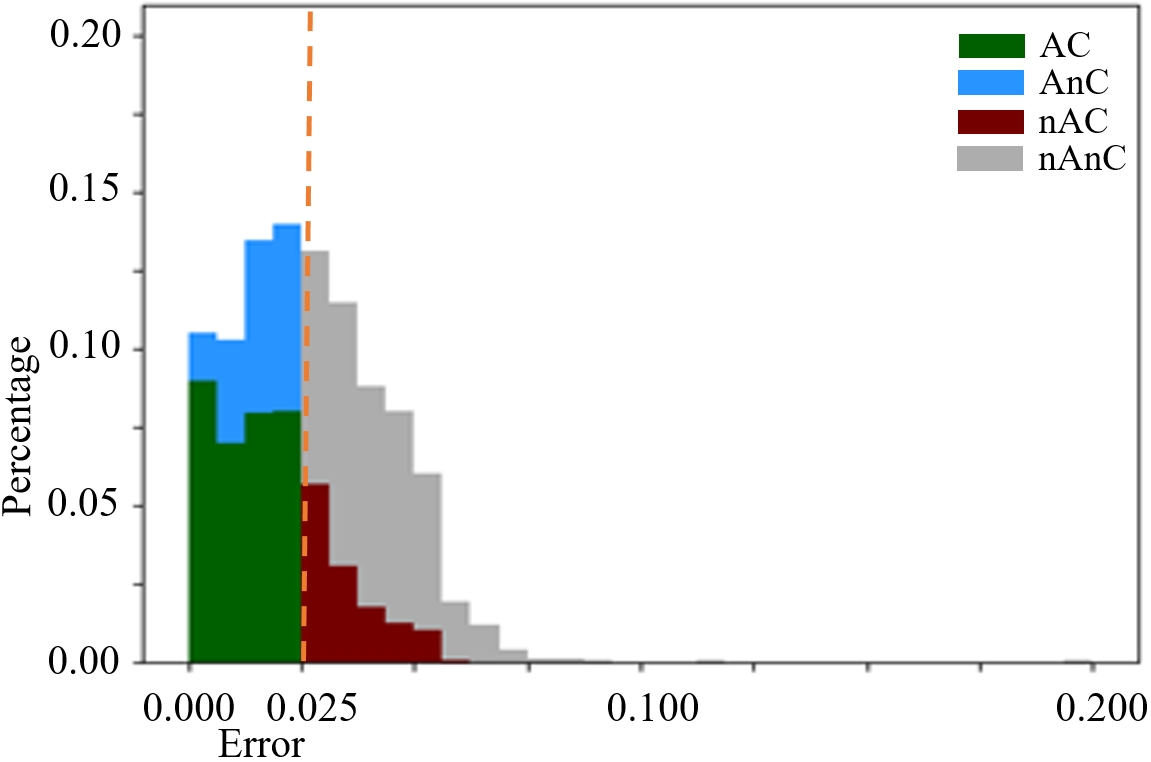}
    }
    \subfigure[Iterative method]{\includegraphics[width=0.3\linewidth]{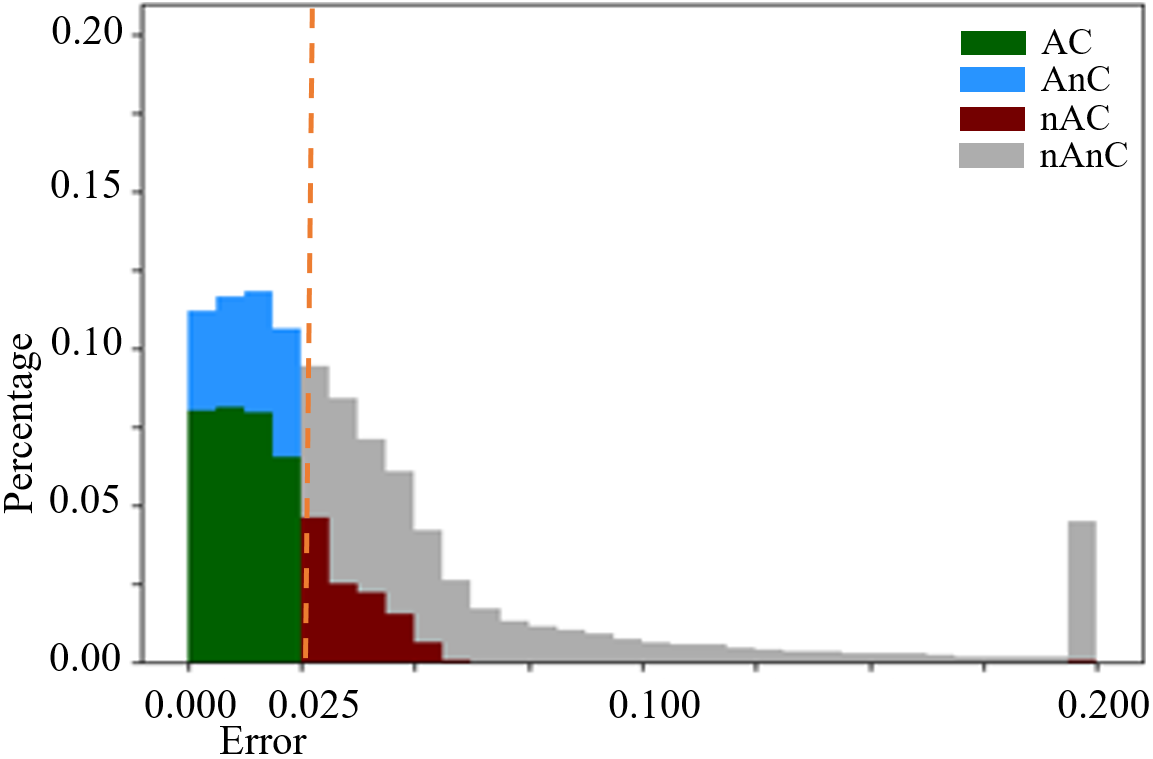}
    }
    \subfigure[MCMA method]{\includegraphics[width=0.3\linewidth]{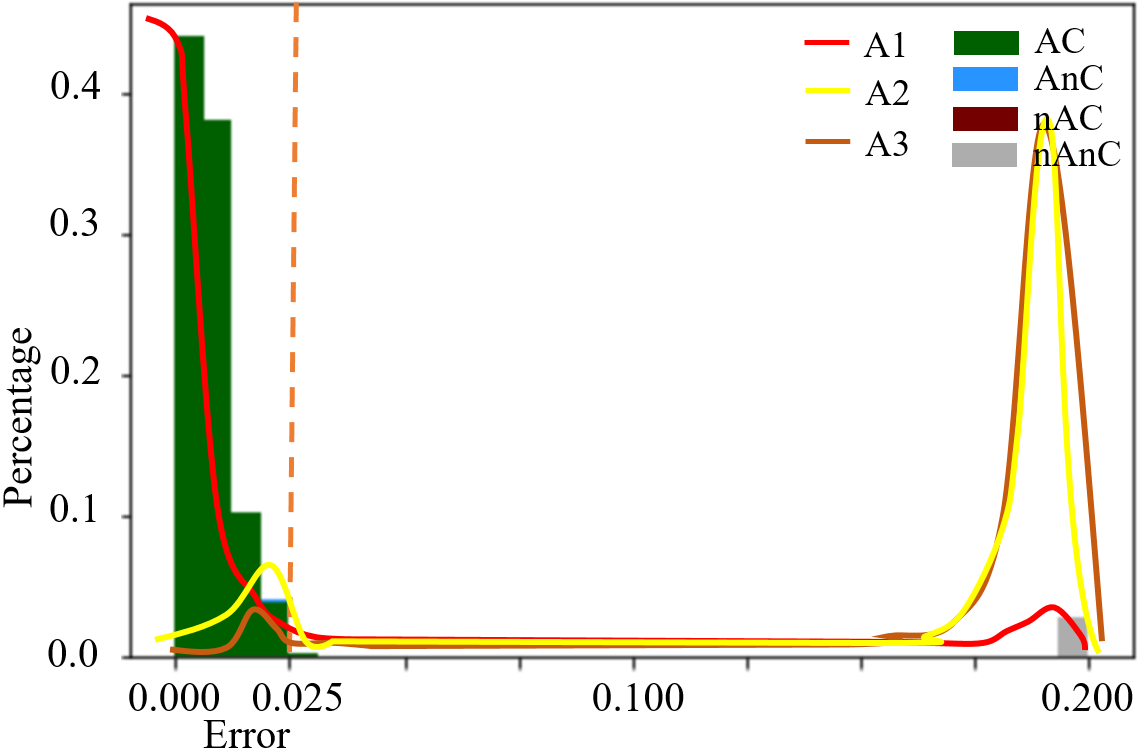}
    }
    \caption{Data distribution of the Bessel bench along the approximation error}
    \label{fig:data_distribution}
    \vspace{-15pt}
\end{figure*}

Fig.~\ref{fig:itr} illustrates the invocation rate for two allocation schemes in the iterative training process of the MCMA architecture using Bessel bench. 
Competitive scheme involves less invocation in the beginning, 
but it keeps progressing in the following iterations and outperforms 
the complementary scheme. We also observe that the invocation of approximator using complementary allocation scheme drops in the second iteration. A possible explanation is that the multi-class classifier begins to work until the second iteration. The classifier shuffles the partition of the training data dramatically and redistributes them to all the approximators. The approximators strive to adapt to this drastic change of training data.  


Because Bessel bench has two-dimensional input, it is easy to show the data distribution with graphs. Without loss of generality, we studied the data distribution of Bessel bench by plotting 
the data samples of our three approximators. We first map these samples to a 2-dimension feature space with two input dimension of Bessel function as the X- and Y- axis, as shown in Fig.~\ref{fig:3dapprx}~(a). Each approximator in the MCMA architecture has its own specialty in fitting a cluster of samples. 
To plot the errors~(Z-axis) of the above data samples, we build a 3-dimension distribution of the samples,  
as shown in Fig.~\ref{fig:3dapprx}~(b). We see that each approximator generates results with a large error in some area. However, with the cooperation of the multi-class classifier, all the approximators approximate a large portion of data samples under the error bound. 


Fig.~\ref{fig:data_distribution} shows the distribution of data samples on approximation error in the \emph{one-pass}, \emph{iterative}, and \emph{MCMA} method. In the legend, A means safe-to-approximate data, C means data predicted acceptable by the classifier. So AC means true positive, nAnC for true negative, AnC means false negative, and nAC for false positive.
In  Fig.~\ref{fig:data_distribution}~(a), after using \emph{one-pass}, many data samples stick around the error bound(dashed line). This phenomenon indicates the hardness to classify the safe-to-approximate data from the others. With \emph{iterative} method, as shown in Fig.~\ref{fig:data_distribution}~(b), the approximator evolves to provide a more accurate output of those accepted data (the green portion); while the classifier becomes more robust to discriminate those unsafe-to-approximate data(fewer samples around the error bound). In Fig.~\ref{fig:data_distribution}~(c), $A_i$ means error distribution curve of the $i_{th}$ approximator. We can see the first approximator covers most of the data and the other two approximators cover the rest part. Our architecture dramatically increases the true invocation rate~(true positive with label \emph{AC}). The MCMA architecture almost recognizes all the safe-to-approximate samples~(low false negative rate), resulting in a high recall of the classifier. In fact, if we draw a classification hyperplane, the hyperplane is almost vertical and approaches to the error bound, indicating a near-optimal classifier.

\section {Conclusion}\label{sect:conclusion}
Neural approximate computing with fine-grain quality control is promising to gain energy-efficiency and performance by the trade-off the tolerable errors in RMS applications. In this paper, we propose new methods to improve the invocation of approximator. Based on the analysis of data distribution, we propose novel approximate computing architectures by orchestrating multiple classifiers and approximators; the corresponding training methods and hardware design are also described. The proposed architectures can dramatically improve the invocation of the approximate accelerator with quality guarantee for larger gain of energy-efficiency. 


\bibliographystyle{unsrt}
\bibliography{nnacc}
\end{document}